\documentclass[runningheads]{llncs}
\usepackage{hyperref}
\usepackage{float}
\usepackage[T1]{fontenc}
\usepackage{amsmath}
\usepackage{graphicx}
\usepackage{amssymb}
\usepackage{multirow}
\usepackage{makecell}
\usepackage{booktabs}

\begin{document}
\title{Stacked LoRA for Subject-Adaptive EEG Foundation Models in Motor Imagery Decoding}
\titlerunning{Stacked LoRA for Subject-Adaptive EEG Decoding}
\author{Aymen Sarhane*\inst{1} \and
Fouad Lbakali*\inst{1} \and
Mouad Souissi*\inst{1} \and
Jonathan Lys\inst{1} \and
Giulia Lioi\inst{1}}
\authorrunning{A. Sarhane et al.}
\institute{IMT Atlantique, Lab-STICC, Brest, France}

\maketitle            
\begin{abstract}
Electroencephalography (EEG) decoding for brain--computer interfaces (BCIs) faces a major challenge: substantial inter-subject variability limits effective cross-subject generalization. Consequently, practical systems still rely largely on subject-specific models trained from scratch and requiring individual recalibration. EEG foundation models have recently emerged as a promising alternative; however, even large pretrained models cannot simply be used as fixed feature extractors and still require additional adaptation before they can be reliably applied to downstream tasks. In this work, we address this challenge through targeted adaptation strategies. Building on recent EEG foundation models such as REVE, LaBraM, and LUNA, we examine the impact of different low-rank adaptation strategies on motor imagery classification. We propose a framework that structurally decouples subject-invariant knowledge from subject-specific neural signatures: the low-rank update at each adapted layer is split into a \emph{Global} adapter, trained jointly across all subjects, and \emph{Subject-Specific} adapters, each absorbing individual variability. To assess the contribution of each path, we compare three adaptation strategies: (i) subject-specific LoRA (ii) global LoRA and (iii) stacked LoRA, combining both Global and Subject Specific adapters. Experiments on BCI Competition IV-2a, PhysioNet Motor Imagery, and the clinical Zuo2025 benchmark show that Stacked LoRA effectively mitigates inter-subject variability, achieving the best accuracy in the large majority of backbone and dataset combinations. Our analysis further reveals that the optimal balance between the global and subject-specific paths depends on the target population: a shared adapter is sufficient for large, diverse cohorts, whereas subject-specific adaptation is decisive under the high inter-session variability of clinical recordings.

\end{abstract}
\section{Introduction}

Electroencephalography (EEG) allows non-invasive brain monitoring, enabling a wide range of clinical and research applications, from epilepsy diagnosis to brain--computer interfaces (BCIs)~\cite{ref_100yearsEEG,ref_bcicompiv}. Among BCI paradigms, motor imagery (MI) decoding has attracted particular attention for its potential to restore communication and motor control in patients with severe neurological impairments~\cite{ref_lioi_multimodal,ref_physionet_bci2000}. However, translating MI decoding into robust real-world systems remains an open challenge, primarily because EEG signals exhibit pronounced inter-subject variability: the neural patterns associated with a given mental state differ substantially across individuals due to differences in anatomy, electrode placement, and cognitive strategy. This variability limits the generalization of models trained on one population to new subjects, and most practical BCI pipelines still rely on costly per-subject recalibration sessions ~\cite{ref_lotte}

Foundation models have recently emerged as a promising paradigm to address this limitation. Trained on large corpora that pool data from many subjects, settings, and tasks, these models learn rich latent representations that can be transferred to downstream applications with minimal additional supervision ~\cite{ref_liu_review}. In the EEG domain, REVE~\cite{ref_reve} is a state-of-the-art foundation model pre-trained on more than 60{,}000 hours of EEG recordings from 25{,}000 subjects via masked autoencoding ~\cite{ref_mae}, ranked first(on average) over a range of 36 EEG tasks in a recent large benchmarking effort ~\cite{ref_neural_bench}. Thanks to its 4D Fourier positional encoding , REVE can natively handle arbitrary electrode layouts and recording lengths, making it directly applicable to heterogeneous datasets collected under different experimental protocols. Yet, even large foundation models cannot simply be deployed as frozen feature extractors: the shift between the general pre-training distribution and the specifics of a given downstream task or subject population demands targeted fine-tuning.  Recent work has tackled this problem from different angles. Klein et al.~\cite{ref_klein} introduced the Subject-Conditioned Layer, which decomposes every adapted weight into a shared component and a low-rank subject-specific correction, demonstrating improvements on a single MI dataset (BCI Competition IV ~\cite{ref_bci2a}) with both CNN and ViT-like architectures. 

Parameter-efficient fine-tuning (PEFT) techniques, and Low-Rank Adaptation (LoRA)~\cite{ref_lora} in particular, have become the method of choice for adapting large foundation models at moderate computational cost. By injecting small trainable rank decomposition matrices into the frozen backbone, LoRA reduces the number of updated parameters from millions to thousands while achieving performance competitive with full fine-tuning. However, naive application of a single shared LoRA to a multi-subject dataset risks conflating two distinct sources of variation: class-discriminative patterns that are genuinely shared across subjects, and subject-specific neural signatures that are specific to each individual. Failing to separate these components means the adapter must simultaneously learn generalizable representations to adapt to the downstream task and account for individual noise, an objective that is inherently conflicting. Recent research has increasingly focused on managing interference between competing objectives in PEFT. MTLoRA~\cite{ref_MTLora} extended low-rank adaptation to multi-task settings by explicitly decomposing parameter updates into task-shared and task-specific components, reducing cross-task interference while preserving a common backbone. Jiang et al.~\cite{ref_Jiang} tackled a complementary challenge, proposing a fine-tuning strategy that shields a reserved majority of parameters from gradient updates, protecting stable pretrained representations from noisy or conflicting task signals during adaptation. Building on these insights, DELoREAN~\cite{ref_delorean} proposed a dual-path adapter architecture that trains a shared adapter alongside disposable task-specific expert adapters for natural language processing; the experts absorb domain-specific noise during training and are discarded at inference, so that only the robust shared representations generalize to unseen data. DELoREAN showed consistent gains over standard LoRA on out-of-distribution benchmarks across five languages and multiple vision-language datasets, and provided theoretical and empirical evidence that the two adapter paths learn orthogonal subspaces.

In this paper, we bring together these lines of work and apply them to the problem of subject-adaptive EEG decoding with foundation models. Building on recent EEG foundation models, we propose \textsc{Stacked LoRA}, an adaptation framework that explicitly decouples task-level knowledge from subject-specific neural signatures within the same forward pass. The weight update is decomposed into (i) a \emph{Global adapter}, trained jointly on all subjects and capturing shared task representations, and (ii) \emph{Subject-Specific adapters}, one per subject, that absorb individual variability without impacting the shared path. We compare three instantiations of this framework (Global LoRA, Subject-Specific LoRA, and Stacked LoRA) against a linear probing baseline, applied as a drop-in adaptation mechanism to three recent EEG foundation models -- REVE ~\cite{ref_reve}, LaBraM~\cite{ref_labram}, and LUNA~\cite{ref_labram}, on three publicly available MI datasets: BCI Competition IV-2a~\cite{ref_bci2a}, PhysioNet Motor Imagery~\cite{ref_physionet}, and the clinical benchmark Zuo2025~\cite{ref_zuo2025}. Our experiments reveal complementary strengths of each strategy and provide practical guidance on when subject-specific specialization is most beneficial.

Our main contributions are: (i) we propose Stacked LoRA, a dual adaptation architecture that decomposes each low-rank update into a Global adapter shared across subjects and per-subject Subject-Specific adapters; (ii) we instantiate it as a drop-in mechanism on three EEG foundation models, REVE, LaBraM, and LUNA, using a single fine-tuning stage on top of the frozen backbone; and (iii) we conduct a systematic evaluation across three MI datasets of varying size, channel count, and clinical relevance, showing that the optimal balance between global and subject-level adaptation depends on the structure of the target population.

\section{Methodology}
\subsection{Problem Formulation}

We consider the multi-subject EEG classification problem.
Let $\mathcal{Y} = \{1, \ldots, C\}$ denote the set of $C$ classes
(e.g.\ left hand, right hand, foot, tongue for MI).
Data are collected from $N$ subjects; the recording of subject $j$
defines a domain $\mathcal{S}_j$, composed of labelled EEG trials
$(x, y)$ with $x \in \mathbb{R}^{E \times T}$ ($E$ electrodes, $T$
time samples) and $y \in \mathcal{Y}$.
Because electrode placements, impedance conditions, and individual
neurophysiology differ across participants, the distributions of the $N$
subjects $\{\mathcal{S}_j\}_{j=1}^{N}$  vary substantially between domains, even for the same class label.

Our goal is to learn, for each subject $j$, a classifier
$f_j : \mathbb{R}^{E \times T} \to \mathcal{Y}$ that minimizes the
expected cross-entropy loss over $\mathcal{S}_j$, while leveraging
the shared structure present across all domains. 
We address this by building $f_j$ on top of a large pre-trained EEG
foundation model and using PEFT to adapt it with a small number of
additional parameters.

\subsection{Low-Rank Adaptation (LoRA)}
Given a pretrained weight matrix $W\in\mathbb{R}^{d_{out}\times d_{in}}$,
LoRA~\cite{ref_lora} adapts the corresponding layer without modifying $W$
itself. The update is parameterized as a rank-$r$ factorization $BA$, with
$B\in\mathbb{R}^{d_{out}\times r}$, $A\in\mathbb{R}^{r\times d_{in}}$, and
$r\ll\min(d_{in},d_{out})$; only $A$ and $B$ receive gradients. The resulting
forward pass is as follows
\begin{equation}
    h = Wx + \frac{\alpha}{r}\, BAx,
\end{equation}
where $\alpha$ is a fixed scaling coefficient. This reduces the per-layer
trainable footprint from $d_{out}\,d_{in}$ to $r(d_{out}+d_{in})$ parameters
while keeping the original backbone untouched.

\subsection{Stacked LoRA}
Our adaptation strategy extends the DELoREAN framework of~\cite{ref_delorean},
which splits a LoRA update between a component shared across all training
domains and one component per domain. DELoREAN was introduced to address
task-level variability in language and vision. Here we repurpose the same
two-component structure for a context in which the source of variance
is the subject. The variant we propose,
\textsc{Stacked LoRA}, dedicates one branch of the low-rank update to
population-level decoding regularities, and parallel branches,
one per subject, to the individual deviations from that common behavior.

On every layer instrumented with adapters, the low-rank update is written as
the sum of a \emph{Global} term, always active, and a \emph{Subject-Specific}
term, active only on trials of the matching subject. For an input $x$ drawn
from subject $j\in\{1,\dots,N\}$, the layer output is
\begin{equation}
    h = \left(
        W
        + \frac{\alpha_g}{r_g}\, B_g A_g
        + \frac{\alpha_s}{r_s} \sum_{k=1}^{N} \mathbf{1}\{x\in\mathcal{S}_k\}\,
        B_k A_k
    \right) x,
\end{equation}
with $(A_g,B_g)$ the Global adapter of rank $r_g$ and scale $\alpha_g$, and
$\{(A_k,B_k)\}_{k=1}^{N}$ the $N$ Subject-Specific adapters, each of rank
$r_s$ and scale $\alpha_s$. The indicator $\mathbf{1}\{x\in\mathcal{S}_k\}$
implements a hard, subject-based routing: for any given trial, exactly one
Subject-Specific adapter is selected, namely the one whose index matches the
trial's origin. The backbone $W$ is kept frozen throughout.

\paragraph{Global adapter.}
Since $(A_g,B_g)$ receives gradients from every trial in the batch, its
optimum depends on the mixture of all subject distributions rather than on any
single subject distribution $\mathcal{S}_j$. It therefore converges toward the portion of the
input–output mapping that the frozen backbone does not already provide and
that is consistent across the whole training set, \emph{i.e.} the class-discriminative component of the EEG signal, following the elimination of subject-specific noise. In this sense, $(A_g,B_g)$ behaves as a
task-level prior over the downstream task.

\paragraph{Subject-Specific adapters.}
Each $(A_j,B_j)$ is trained on trials of subject $j$ only and different subject branches thus receive updates from disjoint slices of every batch. Because the backbone and the Global adapter already model the shared patterns across the group, the per-subject branches do not need to reproduce it, they are left to model the residual that is characteristic of a single participant.

\subsection{Adaptation Strategies}
\label{sec:adaptation_strategies}

\begin{figure}[t]
    \centering
    \includegraphics[width=0.75\linewidth]{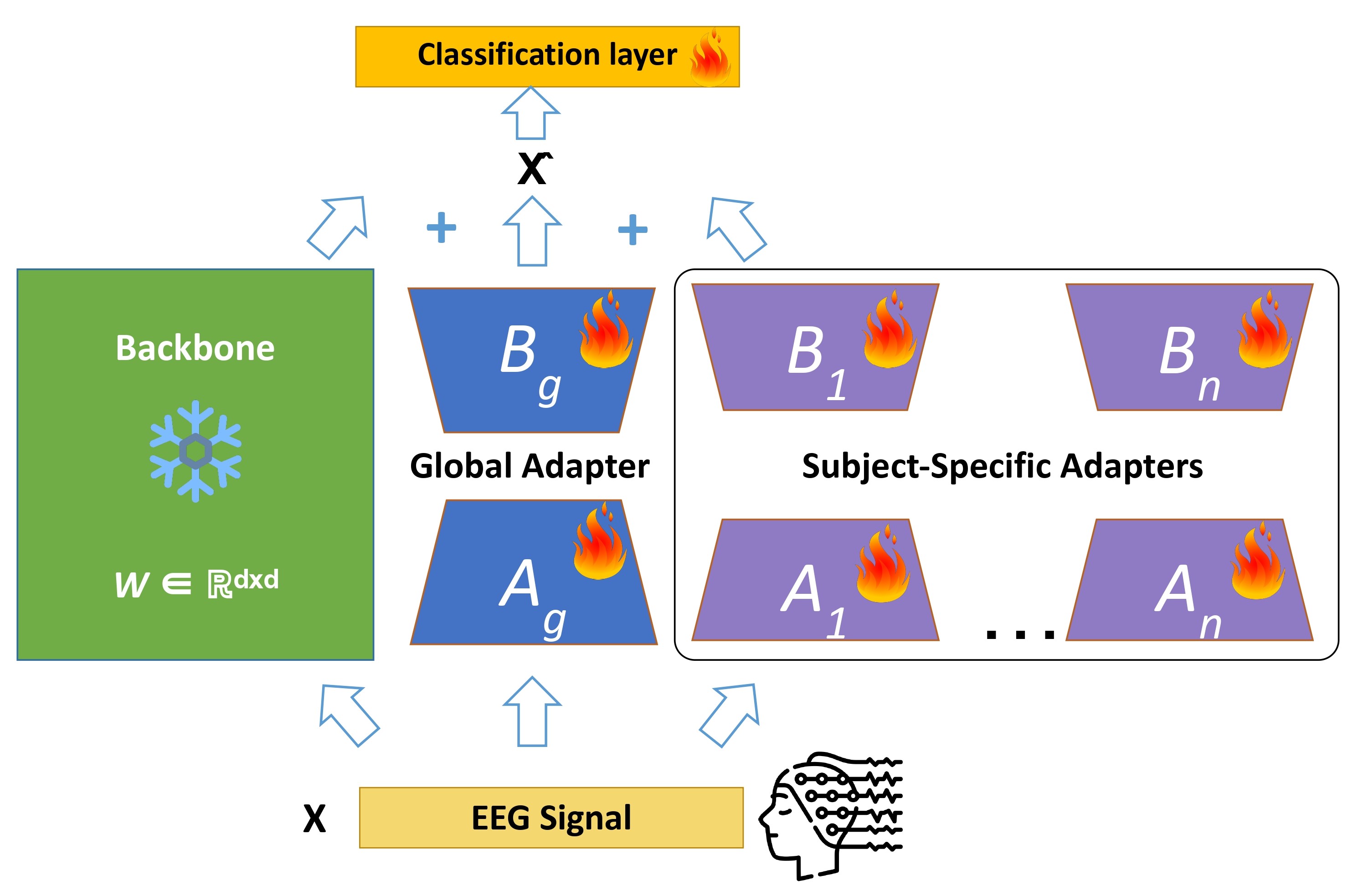}
    \caption{Stacked LoRA framework: the frozen backbone (pretrained EEG foundation model) is augmented with a shared Global Adapter $(A_g, B_g)$ and $N$ Subject-Specific Adapters $(A_k, B_k)$.}
    \label{fig:stacked_lora}
\end{figure}

To isolate the contribution of each path and identify where subject-adaptive
gains actually come from, we instantiate the framework under three
complementary strategies:
\begin{itemize}
  \item \textit{Global LoRA.} Only the Global Adapter is retained, yielding a
  single shared low-rank update trained jointly on all subjects. This strategy
  serves as a population-level baseline that ignores subject identity at
  fine-tuning time.
  \item \textit{Subject-Specific LoRA.} Only the Subject-Specific adapters are
  retained, with one independent low-rank adapter per subject and no shared
  component. This strategy probes the importance of subject-level
  specialization when no cross-subject sharing is enforced.
  \item \textit{Stacked LoRA.} Both paths are active simultaneously, combining
  the Global Adapter with the Subject-Specific adapters within the same
  forward pass, this is the full framework.
\end{itemize}

\subsection{Training Details}
\label{sec:training}

All strategies share a single fine-tuning stage on top of a frozen backbone.
A classification head is appended on top of the frozen backbone: the output tokens are first flattened, normalized with
RMSNorm~\cite{ref_rmsnorm}, passed through a dropout layer
($p = 0.1$), and projected to the number of target classes by a single
linear layer. Training uses AdamW~\cite{ref_adamw}
with a learning rate of $10^{-4}$ and cosine-annealing
with $10\%$ linear warmup, for up to $25$ epochs with early stopping on the
validation accuracy (patience $= 5$). The LoRA rank is set
to $r = 32$ for the Global adapter and $r = 8$
for the Subject-Specific adapters; all other optimisation choices are shared
across strategies.

\section{EEG Foundation Models}

To demonstrate the generality of Stacked LoRA as a drop-in adaptation mechanism, we benchmark
it on three EEG foundation models: REVE \cite{ref_reve}, LaBraM \cite{ref_labram}, and
LUNA \cite{ref_luna}. REVE is a large-scale self-supervised foundation model pretrained on
over 60{,}000 hours of EEG; it segments each channel into temporal patches, encodes them with
a 4D Fourier-based positional encoding that jointly captures 3D electrode coordinates and patch
index, and processes them with a spatio-temporal transformer encoder, enabling it to handle
arbitrary electrode layouts natively. LaBraM \cite{ref_labram} is a neural Transformer
pretrained on approximately 2{,}500 hours of EEG from around 20 datasets using
vector-quantized masked patch prediction. LUNA \cite{ref_luna} is a topology-agnostic
foundation model that projects variable-channel EEG into a fixed-size latent space via learned
queries and cross-attention before applying temporal self-attention, pretrained on over 21{,}000
hours across diverse electrode montages. We integrate Stacked LoRA by applying its Global and
Subject-Specific adapters to the linear projections within the transformer encoder blocks of each
model, namely the query, key, value, and feed-forward weight matrices.

\section{Evaluation Datasets}
We conduct our experiments on three Brain Computer Interface datasets, all publicly available through the MOABB package~\cite{ref_moabb}~\cite{ref_moabb_benchmark}, and none of which were part of REVE's pre-training data. The BCI Competition IV 2a dataset~\cite{ref_bcicompiv}~\cite{ref_bci2a} consists of 22-channel EEG recordings from nine healthy subjects. For each subject, data were collected in two sessions (training and evaluation), with 288 trials per session and a 4-second MI window per trial, yielding approximately 38 minutes of labelled EEG per subject and 5.8 hours in total. The labels are balanced across four MI classes: left hand, right hand, foot, and tongue. The PhysioNet dataset~\cite{ref_physionet} consists of 64-channel EEG recordings from 109 healthy subjects. Each subject performed 6 runs of MI tasks with a 3-second window per trial, yielding approximately 9 minutes per subject and over 16 hours in total; following common practice we use the four-class setup: left fist, right fist, both fists, and both feet. Finally, the Zuo2025 dataset~\cite{ref_zuo2025} consists of 30-channel EEG recordings from 30 subjects, collected across 5 sessions, with 4 trials per session and a 4-second window per trial, yielding approximately 1.3 minutes per subject and 40 minutes in total. The labels are balanced across 2 MI classes: left leg and right leg. Unlike the two former datasets, which involve healthy subjects, Zuo2025 was recorded on patients, providing a clinically relevant benchmark to assess the generalization of our approach beyond healthy populations.

\paragraph{Preprocessing and Data Splitting.}
Each dataset is preprocessed according to the input statistics expected by the respective pretrained checkpoint. For REVE, signals are resampled to 200\,Hz, bandpass-filtered between 0.5 and 99.5\,Hz, standardized channel-wise using a standard scaler, and clamped at 15 standard deviations. For LaBraM, signals are resampled to 200\,Hz, bandpass-filtered between 0.1 and 75\,Hz, with a notch filter at 50\,Hz, and rescaled by a fixed factor of $10^{-4}$ to convert from volts to 0.1\,mV units. For LUNA, signals are resampled to 256\,Hz, bandpass-filtered between 0.1 and 75\,Hz, with notch filters at 50 and 60\,Hz; per-trial channel-wise z-score normalization is applied internally by the model. All datasets are split into training, validation, and test sets using a random stratified partition. To evaluate our two-stage fine-tuning framework while strictly preventing data leakage, we perform a single global split. Trials from all subjects are pooled and randomly partitioned into training (70\%), validation (15\%), and test (15\%) sets. The initial global fine-tuning stage utilizes this pooled training set. For the subsequent subject-specific adaptation, the data is not re-split; instead, subject-level loaders are constructed by filtering the original global partitions. This structural design guarantees that every subject-specific test set is strictly disjoint from the pooled training data.

\section{Results}
\label{sec:results}
 
We evaluate the different adaptation strategies across three EEG foundation models and three MI benchmarks that span a wide range of cohort sizes, channel counts, and
clinical conditions, demonstrating the generality of our approach.
Table~\ref{tab:main_results} reports the average cross-subject balanced accuracy
for every backbone and strategy. Across the nine backbone--dataset combinations,
Stacked LoRA attains the best or tied-best balanced accuracy in seven; the two
exceptions are PhysioNet MI with LaBraM and LUNA, where the Global path alone is
already optimal.
 
\begin{table}[ht]
\centering
\caption{Average cross-subject balanced accuracy (\%, mean $\pm$ standard 
error of the mean across 3 seeds).}
\label{tab:main_results}
\begin{tabular}{llccc}
\hline
\textbf{Model} & \textbf{Strategy} & \textbf{BCI IV-2a} & \textbf{PhysioNet MI} & \textbf{Zuo2025}\\
 & & \textbf{(9 subj.)} & \textbf{(109 subj.)} & \textbf{(30 subj.)}\\
\hline
\multirow{4}{*}{REVE}
 & Linear Probing         & $61.7 \pm 0.9$ & $61.0 \pm 0.4$ & $87.0 \pm 0.1$ \\
 & Global LoRA            & $76.4 \pm 0.3$ & $66.7 \pm 0.8$ & $90.8 \pm 0.3$ \\
 & Subject-Specific LoRA  & $76.0 \pm 1.0$ & $65.5 \pm 0.4$ & $\mathbf{92.9 \pm 0.0}$ \\
 & Stacked LoRA           & $\mathbf{80.3 \pm 0.9}$ & $\mathbf{67.8 \pm 0.5}$ & $92.9 \pm 0.2$ \\
\hline
\multirow{4}{*}{LaBraM}
 & Linear Probing         & $44.1 \pm 0.9$ & $49.1 \pm 0.5$ & $77.2 \pm 0.3$ \\
 & Global LoRA            & $56.8 \pm 1.0$ & $\mathbf{60.2 \pm 0.8}$ & $84.9 \pm 0.2$ \\
 & Subject-Specific LoRA  & $54.6 \pm 0.7$ & $53.3 \pm 0.9$ & $83.2 \pm 0.2$ \\
 & Stacked LoRA           & $\mathbf{61.3 \pm 0.4}$ & $59.9 \pm 0.7$ & $\mathbf{86.1 \pm 0.1}$ \\
\hline
\multirow{4}{*}{LUNA}
 & Linear Probing         & $40.9 \pm 1.1$ & $50.7 \pm 0.1$ & $79.4 \pm 0.4$ \\
 & Global LoRA            & $47.6 \pm 1.0$ & $\mathbf{55.1 \pm 0.3}$ & $84.5 \pm 0.7$ \\
 & Subject-Specific LoRA  & $45.6 \pm 0.5$ & $44.8 \pm 0.5$ & $86.8 \pm 0.5$ \\
 & Stacked LoRA           & $\mathbf{49.7 \pm 1.6}$ & $51.9 \pm 0.8$ & $\mathbf{87.2 \pm 0.2}$ \\
\hline
\end{tabular}
\end{table}
 
\subsection{Classification Performance}
 
We first ask whether targeted adaptation is needed at all. Across all three
foundation models, every LoRA-based strategy improves substantially over linear
probing, establishing that adaptation is a prerequisite before these models can
be applied to MI decoding. The effect is most pronounced with REVE on
BCI IV-2a, where Stacked LoRA improves over linear probing by $18.6$ points
($80.3\%$ vs.\ $61.7\%$), and it persists for LaBraM ($+17.2$ points) and LUNA
($+8.8$ points). Even the strongest pretrained EEG backbone therefore cannot be
deployed as a frozen feature extractor.
 
We next examine the two benchmarks in which subject-level adaptation proves
valuable, namely BCI IV-2a and the clinical Zuo2025 dataset. In both, Stacked
LoRA reaches the best or tied-best accuracy for every backbone. On BCI IV-2a it
improves over Global LoRA by $3.9$ points and over Subject-Specific LoRA by $4.3$
points with REVE ($80.3\%$), with the same ordering for LaBraM ($61.3\%$) and
LUNA ($49.7\%$): the two paths are genuinely complementary, combining shared
task-discriminative representations with individual deviations. On Zuo2025 the
gains are carried mainly by the subject-specific path, Subject-Specific LoRA
ties Stacked LoRA for REVE ($92.9\%$ for both) and trails it only marginally for
LUNA ($86.8\%$ vs.\ $87.2\%$), while Stacked LoRA stays best for LaBraM
($86.1\%$), reflecting the high inter-session variability of patient
recordings, for which per-subject adaptation is particularly effective. As shown
These improvements are distributed across
subjects rather than driven by a few individuals.
 
The picture changes only when the subject pool becomes large and diverse. On the
109-subject PhysioNet benchmark, Global LoRA matches or exceeds Subject-Specific
LoRA for every backbone, marginally for REVE ($66.7\%$ vs.\ $65.5\%$) and
substantially for LaBraM ($60.2\%$ vs.\ $53.3\%$) and LUNA ($55.1\%$ vs.\
$44.8\%$). Stacked LoRA remains best for REVE ($67.8\%$), but for LaBraM and LUNA
the Global path alone is already optimal ($60.2\%$ and $55.1\%$), with the
additional subject-specific branches bringing no further gain. With a large and
diverse subject pool, the shared adapter captures enough cross-subject structure
to generalise, whereas per-subject adapters, each trained on relatively few
trials, offer diminishing returns.

\subsection{Parameter Efficiency and Deployment Cost}
 
Although Stacked LoRA trains $S$ subject-specific adapters jointly, the model
actually deployed for a given subject combines only the classification head, the
Global adapter, and that subject's rank-8 adapter, and therefore remains compact
(Table~\ref{tab:params}). For REVE, the per-subject Stacked model activates
$9.7\%$ of the full parameters ($7.42$M), against $8.0\%$ for Global LoRA and
$2.3\%$ for Subject-Specific LoRA alone. The higher training cost of the full
framework thus buys the best accuracy without substantially inflating the number of parameters.
 
\begin{table}[ht]
\centering
\small
\caption{Trainable parameters per fine-tuning strategy: share of the 
full model (trainable\,/\,(frozen backbone $+$ trainable)), with the 
count in millions in parentheses. Frozen backbones: REVE 69.2M, LaBraM 
5.8M, LUNA 40.4M. $S$ is the number of subjects (BCI IV-2a 9, PhysioNet 
MI 109, Zuo2025 30); $\text{LoRA}_{r}$ is a rank-$r$ adapter on all 
targeted linear layers. For Subject-Specific and Stacked LoRA the upper 
value is the \emph{total} optimised over all $S$ subjects and the lower 
value the \emph{per-subject} deployed model (head $+$ global adapter $+$ 
one subject's rank-8 adapter).}
\label{tab:params}
\resizebox{\textwidth}{!}{%
\begin{tabular}{lllccc}
\toprule
\textbf{Model} & \textbf{Strategy} & \textbf{Trained params} & \textbf{BCI IV-2a} & \textbf{PhysioNet MI} & \textbf{Zuo2025}\\
\midrule
\multirow{6}{*}{REVE}
 & Linear Probing        & head only                                            & $0.3\%$ ($0.23\,\text{M}$) & $0.7\%$ ($0.49\,\text{M}$) & $0.3\%$ ($0.18\,\text{M}$) \\
 & Global LoRA           & head $+ \text{LoRA}_{32}$                            & $8.0\%$ ($5.98\,\text{M}$) & $8.3\%$ ($6.25\,\text{M}$) & $7.9\%$ ($5.94\,\text{M}$) \\
\cmidrule{2-6}
 & \multirow{2}{*}{Subject-Specific LoRA} & head $+ S \times \text{LoRA}_{8}$ \hfill (total) & $16.0\%$ ($13.18\,\text{M}$) & $69.5\%$ ($157.40\,\text{M}$) & $38.5\%$ ($43.37\,\text{M}$) \\
 &                        & head $+ \text{LoRA}_{8}$ \hfill (per-subject)        & $2.3\%$ ($1.66\,\text{M}$)  & $2.7\%$ ($1.93\,\text{M}$)   & $2.3\%$ ($1.62\,\text{M}$) \\
\cmidrule{2-6}
 & \multirow{2}{*}{Stacked LoRA}          & head $+ \text{LoRA}_{32} + S \times \text{LoRA}_{8}$ \hfill (total) & $21.5\%$ ($18.94\,\text{M}$) & $70.2\%$ ($163.16\,\text{M}$) & $41.5\%$ ($49.13\,\text{M}$) \\
 &                        & head $+ \text{LoRA}_{32} + \text{LoRA}_{8}$ \hfill (per-subject)    & $9.7\%$ ($7.42\,\text{M}$)   & $10.0\%$ ($7.69\,\text{M}$)  & $9.6\%$ ($7.38\,\text{M}$) \\
\midrule
\multirow{6}{*}{LaBraM}
 & Linear Probing        & head only                                            & $8.1\%$ ($0.51\,\text{M}$)  & $6.2\%$ ($0.38\,\text{M}$)  & $5.0\%$ ($0.31\,\text{M}$) \\
 & Global LoRA           & head $+ \text{LoRA}_{32}$                            & $23.0\%$ ($1.74\,\text{M}$) & $21.7\%$ ($1.61\,\text{M}$) & $20.9\%$ ($1.54\,\text{M}$) \\
\cmidrule{2-6}
 & \multirow{2}{*}{Subject-Specific LoRA} & head $+ S \times \text{LoRA}_{8}$ \hfill (total) & $36.0\%$ ($3.28\,\text{M}$)  & $85.3\%$ ($33.87\,\text{M}$) & $62.1\%$ ($9.52\,\text{M}$) \\
 &                        & head $+ \text{LoRA}_{8}$ \hfill (per-subject)        & $12.3\%$ ($0.82\,\text{M}$) & $10.6\%$ ($0.69\,\text{M}$)  & $9.6\%$ ($0.61\,\text{M}$) \\
\cmidrule{2-6}
 & \multirow{2}{*}{Stacked LoRA}          & head $+ \text{LoRA}_{32} + S \times \text{LoRA}_{8}$ \hfill (total) & $43.6\%$ ($4.51\,\text{M}$)  & $85.8\%$ ($35.10\,\text{M}$) & $64.9\%$ ($10.75\,\text{M}$) \\
 &                        & head $+ \text{LoRA}_{32} + \text{LoRA}_{8}$ \hfill (per-subject)    & $26.0\%$ ($2.05\,\text{M}$)  & $24.8\%$ ($1.92\,\text{M}$)  & $24.1\%$ ($1.84\,\text{M}$) \\
\midrule
\multirow{6}{*}{LUNA}
 & Linear Probing        & head only                                            & $0.2\%$ ($0.07\,\text{M}$)  & $0.1\%$ ($0.05\,\text{M}$)  & $0.1\%$ ($0.04\,\text{M}$) \\
 & Global LoRA           & head $+ \text{LoRA}_{32}$                            & $6.9\%$ ($3.02\,\text{M}$)  & $6.9\%$ ($3.00\,\text{M}$)  & $6.9\%$ ($2.99\,\text{M}$) \\
\cmidrule{2-6}
 & \multirow{2}{*}{Subject-Specific LoRA} & head $+ S \times \text{LoRA}_{8}$ \hfill (total) & $14.2\%$ ($6.71\,\text{M}$)  & $66.5\%$ ($80.42\,\text{M}$) & $35.4\%$ ($22.16\,\text{M}$) \\
 &                        & head $+ \text{LoRA}_{8}$ \hfill (per-subject)        & $2.0\%$ ($0.81\,\text{M}$)   & $1.9\%$ ($0.79\,\text{M}$)   & $1.9\%$ ($0.78\,\text{M}$) \\
\cmidrule{2-6}
 & \multirow{2}{*}{Stacked LoRA}          & head $+ \text{LoRA}_{32} + S \times \text{LoRA}_{8}$ \hfill (total) & $19.3\%$ ($9.66\,\text{M}$)  & $67.3\%$ ($83.37\,\text{M}$) & $38.3\%$ ($25.11\,\text{M}$) \\
 &                        & head $+ \text{LoRA}_{32} + \text{LoRA}_{8}$ \hfill (per-subject)    & $8.5\%$ ($3.76\,\text{M}$)   & $8.5\%$ ($3.74\,\text{M}$)   & $8.4\%$ ($3.73\,\text{M}$) \\
\bottomrule
\end{tabular}%
}
\end{table}

\section{Discussion}
 
Taken together, our experiments show that no single adaptation strategy dominates
across all settings: the most effective configuration depends on the size and the
variability of the target population, and our framework spans this spectrum within
a single architecture. Three regimes emerge. When subjects are numerous and
diverse, as in PhysioNet MI, a shared Global adapter already captures the
task-relevant signal common to the population, and the per-subject branches, each
trained on few trials, add little. At the opposite end, when inter-session
variability is high and data per subject are scarce, as in the clinical Zuo2025
recordings, subject-specific specialization becomes the primary driver and the
Global path contributes only marginally. BCI IV-2a sits between these extremes,
nine subjects, balanced sessions, moderate variability, and it is precisely
there that activating both paths jointly pays off most, with Stacked LoRA
improving over each path alone by a clear margin. 
 
A second observation cuts across all three regimes. The gap between linear probing
and any of our fine-tuning strategies is large, reaching $18.6$ points with REVE
on BCI IV-2a and staying substantial across the other backbones and datasets, which confirms that even the most capable EEG foundation models cannot be used as
frozen feature extractors. Some degree of targeted adaptation is always required,
and Stacked LoRA provides a principled way to supply it without having to choose,
a priori, between population-level generalization and per-subject personalization.

\section{Limitations}
 
By design, Stacked LoRA relies on subject identity to route each trial to its
Subject-Specific adapter at both training and inference time. A subject not seen
during training therefore cannot benefit from the subject-specific path without
first fitting a new adapter, which precludes true zero-shot personalization;
relaxing the hard subject routing toward adapter sharing or transfer across
similar subjects is left for future work. The training cost also grows with the
number of subjects, since the $S$ Subject-Specific adapters are optimised jointly, on the 109-subject PhysioNet benchmark the total trainable share reaches
$70.2\%$ of REVE, even though the per-subject deployed model stays compact.
Relatedly, the benefit of the subject-specific path depends on having enough
trials per subject: when data per subject are scarce, as on PhysioNet, these
adapters offer diminishing returns. Finally, the adapter ranks and scaling factors
are fixed across datasets and backbones rather than tuned per setting, and our
evaluation is restricted to MI decoding with within-dataset splits;
validating Stacked LoRA on other paradigms and under cross-session protocols
remains an avenue for future work.

\section{Conclusion}
 
We introduced Stacked LoRA, a parameter-efficient adaptation framework that
explicitly decouples shared task knowledge from individual neural signatures for
EEG MI decoding. Across three foundation model backbones and three
datasets, Stacked LoRA matches or outperforms the best single-path strategy in the
large majority of settings while keeping the per-subject deployed model compact.
Beyond raw accuracy, our analysis clarifies when each form of adaptation is most
beneficial: a shared adapter is sufficient for large and diverse cohorts,
subject-specific adapters are decisive under high inter-session variability, and
their combination is most valuable in the intermediate regime, offering
practical guidance for BCI practitioners. More broadly, because subject identity
is only one instance of a known, discrete source of distribution shift, the same
approach may transfer to other BCI decoding problems in
which population-level structure and source-specific deviations must be modelled
jointly.


\end{document}